\documentclass[conference]{IEEEtran}
\IEEEoverridecommandlockouts
\usepackage{cite}
\usepackage{amsmath,amssymb,amsfonts}
\usepackage{algorithmic}
\usepackage{graphicx}
\usepackage{textcomp}
\usepackage{xcolor}
\usepackage{tikz}
\usepackage[algo2e,ruled,linesnumbered,boxed,commentsnumbered]{algorithm2e}
\usepackage{latexsym}
\usepackage{dcolumn}
\usepackage{booktabs}
\usepackage{amsmath}
\usepackage{multirow}
\usepackage{graphicx}
\usepackage{amssymb}
\usepackage{pifont}
\usepackage{amsmath,array,graphicx}
\usepackage{kantlipsum}

\def\BibTeX{{\rm B\kern-.05em{\sc i\kern-.025em b}\kern-.08em
    T\kern-.1667em\lower.7ex\hbox{E}\kern-.125emX}}
   
\begin{document}

\title{Uncovering Promises and Challenges of Federated Learning to Detect Cardiovascular Diseases: A Scoping Literature Review }

\author{
    \IEEEauthorblockN{Sricharan Donkada\IEEEauthorrefmark{1},  Seyedamin Pouriyeh\IEEEauthorrefmark{1}, Reza M. Parizi\IEEEauthorrefmark{2}, Meng Han \IEEEauthorrefmark{3}, 
     Nasrin Dehbozorgi\IEEEauthorrefmark{4}, \\
     Nazmus Sakib\IEEEauthorrefmark{1}, 
     Quan Z. Sheng\IEEEauthorrefmark{5}
     }
     \IEEEauthorblockA{\IEEEauthorrefmark{1} Department of Information and Technology, Kennesaw State University, Marietta, GA, USA
    \\ \ sdonkada@students.kennesaw.edu, \{spouriye, nsakib1\}@kennesaw.edu}
     \IEEEauthorblockA{\IEEEauthorrefmark{2}Decentralized Science Lab, Kennesaw State University, Marietta, GA, USA   \\ rparizi1@kennesaw.edu}
     \IEEEauthorblockA{\IEEEauthorrefmark{3} Binjiang Institute of Zhejiang University, Hangzhou, Zhejiang, China \\ mhan@zju.edu.cn}
     \IEEEauthorblockA{\IEEEauthorrefmark{4} Department of Software and Game Development, Kennesaw State University, Marietta, GA, USA
    \\ \ dehbozorgi@kennesaw.edu}
    \IEEEauthorblockA{\IEEEauthorrefmark{5} School of Computing, Macquarie University, Sydney, Australia \\ michael.sheng@mq.edu.au}

}

\maketitle
\begin{abstract}
Cardiovascular diseases (CVD) are the leading cause of death globally, and early detection can significantly improve outcomes for patients. Machine learning (ML) models can help diagnose CVDs early, but their performance is limited by the data available for model training. Privacy concerns in healthcare make it harder to acquire data to train accurate ML models. Federated learning (FL) is an emerging approach to machine learning that allows models to be trained on data from multiple sources without compromising the privacy of the individual data owners. This survey paper provides an overview of the current state-of-the-art in FL for CVD detection. We review the different FL models proposed in various papers and discuss their advantages and challenges. We also compare FL with traditional centralized learning approaches and highlight the differences in terms of model accuracy, privacy, and data distribution handling capacity. Finally, we provide a critical analysis of FL's current challenges and limitations for CVD detection and discuss potential avenues for future research. Overall, this survey paper aims to provide a comprehensive overview of the current state-of-the-art in FL for CVD detection and to highlight its potential for improving the accuracy and privacy of CVD detection models.
\end{abstract}


\section{Introduction}

Cardiovascular diseases (CVDs) are one of the major public health issues that are causing approximately 18 million deaths a year worldwide\cite{artificial_bee_colony_based_hdd}. CVDs are responsible for a significant proportion of preventable hospitalizations accounting for 30\% (equal to \$9 billion) in the United States\cite{heart_sound_classification}. 
Early detection of CVD can minimize the risk of death and improves the chances of successful treatment\cite{early_detection_ref}. Among various early CVD detection approaches, using machine learning techniques have shown promising results in recent years\cite{deep_learning_for_hd_survey, ml_based_cvd_survey}. 
Machine learning models can be used as accurate and efficient diagnostic tools to identify individuals at an early stage of CVDs, even before the symptoms develop, and can improve their chances of recovery and also reduce healthcare costs\cite{async_ecg_arrhythmia}. Building accurate and unbiased machine learning models, in general, requires a lot of data to train the models which is hard to acquire especially in the healthcare domain due to privacy concerns and regulations such as General Data Protection Regulation (GDPR), and Health Insurance Portability and Accountability Act (HIPAA)\cite{hipaa_gdpr_ref}. Data privacy has become more important than ever before, with tightened rules such as California Consumer Privacy Act (CCPA), GDPR, and HIPAA making it harder to collect data for training centralized machine learning models\cite{non_iid_heart_rate_var}. Recently, Federated Machine Learning (FL) was proposed by Google to address the challenges surrounding privacy and data security\cite{google_fl_communication_ref}. FL is a machine learning technique that allows multiple decentralized devices or systems to collaboratively train a model while keeping their data private and secure\cite{fl_privacy_ref}. FL addresses the challenges of data collection and privacy compliance by enabling the training of a model without the need to share the raw data\cite{fl_param_transfer_ref}. It also helps to protect the privacy of individuals by not sharing or storing their data in a central location\cite{fl_privacy_ref}.

FL is particularly useful in healthcare because of its privacy-preserving feature\cite{fl_in_healthcare_ref}. In the healthcare industry, FL can be used to train predictive models using data from various sources, such as electronic health records (EHRs)\cite{predictive_models_from_electornic_health_records}, biomedical devices, and imaging data\cite{multi_center_imaging_diagnostics}. This approach allows medical institutions to leverage the data they have collected to train the models while maintaining patient privacy and complying with privacy regulations.

Over recent years, FL has shown promising results by training machine learning models without transferring the raw data and preserving privacy. There were various models proposed to detect CVDs using FL. The authors of \cite{chf_based_on_unetpp} proposed an FL-based U-shaped network that classifies patients with congestive heart failure based on the readings from an electrocardiogram. The proposed model has shown better accuracy than the existing models. In another study, the authors of \cite{artificial_bee_colony_based_hdd} proposed an architecture with Federated Matched Averaging as the FL algorithm and a Modified Artificial Bee Colony optimizer for diagnosing heart diseases. This model performed better than existing FL architectures by using less communication bandwidth and higher accuracy. In another effort, Wanyong et al. \cite{heart_sound_classification} compared FedAvg multilayer perceptron and FedAvg convolutional neural networks with their centralized counterparts. The experiment used non-IID data which is close to a real-world scenario and the results show that the FL algorithms performed better than the centralized algorithms. These experiments show the superiority of FL-based algorithms in a decentralized healthcare setup.

In this paper, we reviewed different FL models proposed in various papers and addressed several research questions related to the use of federated learning (FL) in detecting cardiovascular diseases (CVD), including the types of data used in FL for CVD detection, the different FL models proposed in the literature for CVD detection, a comparison of FL performance against traditional machine learning algorithms in CVD detection, and an examination of the challenges and limitations of FL in CVD detection and potential solutions to address them.

The remainder of the paper is organized as follows. Section \ref{LitReview} covers the review of the existing literature. In section \ref{rqs}, we addressed various research questions in this domain. Finally, we highlighted the findings, outlined future work, and concluded in Section \ref{discussion}.

\section{Literature Survey}
\label{LitReview}

Machine learning techniques have been widely used in detecting and diagnosing cardiovascular diseases (CVDs) due to its ability to analyze large amounts of data from various sources such as ECG, PCG, and echocardiography\cite{deep_learning_for_hd_survey, ml_based_cvd_survey}. It enables the training of models on large datasets to improve the accuracy of CVD detection, while also overcoming the limitations of traditional clinical decision-making by analyzing a broader range of data and factors. This results in more accurate and personalized diagnoses, and offers the potential for more effective treatment plans. In this section, we discussed existing survey papers on traditional ML-based heart disease diagnosis.

Brites et al. \cite{brites_ml_iot_survey} conducted a literature review on machine learning using IoT for diagnosing heart diseases using heart sounds. The study was set to answer research questions based on 58 papers on this research topic. Various components of models that use IoT devices for monitoring vital signs such as heart rate, body temperature, blood pressure, and blood oxygenation were presented. The authors highlighted the rapid progress in the field of IoT, also known as the Internet of Health Things (IoHT). The paper also mentioned specific studies that have implemented integrated system solutions for acquiring and analyzing cardiac sound signals with machine learning algorithms. The emphasis was on the importance of safe processes for the use of IoT devices in healthcare and the use of digital stethoscopes in most projects. The authors mentioned the potential for future projects to incorporate internet connectivity into smart stethoscope designs to reduce production costs. The paper also discussed the use of big data, machine learning algorithms, and statistical methods in detecting and diagnosing cardiovascular diseases. The research included the use of deep learning (DL) algorithms and CNN architectures for the automatic detection of cardiac sounds, monitoring heart rate through low-cost wearable devices and statistical algorithms of linear regression, and using ML algorithms such as Extreme Gradient Boosting (XGBoost), Logistic Regression, Decision Tree, k-Nearest Neighbors (KNN), Random Forest, and Support Vector Machine (SVM) for the development of CVD risk scores and classification of cardiac sounds. The studies also explored the use of PCG, a non-invasive method for heart rate classification, and the development of tools to assist clinical decision-making based on machine learning with the XGBoost algorithm.

Cameron et al. \cite{cameron_clinical_apps_of_ml} conducted a survey on clinical applications of machine learning for the diagnosis and prediction of heart failure. The authors discussed that ML has been used in medicine to predict patient outcomes and improve clinical decision-making. However, there are still challenges to overcome before ML algorithms become a standard practice in healthcare. For example, in heart failure, ML algorithms improved prediction capability in some instances but failed to provide better performance over more traditional methods such as logistic regression (LR). In other studies\cite{dl_based_cvd_ref1, dl_based_cvd_ref2, dl_based_cvd_ref3}, deep learning (DL) models modestly improved performance, but their complexity led to a loss of interpretability in prediction results. These results may be a reflection of the difficulty in obtaining high-quality data as well as the natural heterogeneity that exists between patients. As such, these problems may be best represented by simpler models with more inherent interpretability. There are several obstacles remaining before ML algorithms become a part of standard practice. These obstacles include improving data collection and algorithm development, improved interpretability and interoperability, and determining appropriate roles for these algorithms in clinical practice. Additionally, several criteria need to be met before ML algorithms can be widely utilized in clinical practice, including validation on multiple external cohorts, wide applicability to different hospital systems, settings, and geographic locations, a balance between interpretability and model complexity, and accuracy, monitoring to ensure accurate results given new user-driven data, and cost-effectiveness in development and widespread implementation. Clinicians must also ensure proper use and application of ML algorithms, considering questions such as whether the application will be better served by ML as opposed to traditional statistical models, the level of complexity required, which features or variables to include, which outcomes matter most, and how the model will be adapted into clinical practice while gaining the trust of clinicians who will be using it when the model may not be interpretable. Cameron et al. \cite{cameron_clinical_apps_of_ml} concluded that if these obstacles and criteria can be met and the algorithms have been rigorously proven, the benefits of applying AI to medicine may be soon realized.

To the best of our knowledge, this survey paper is the first comprehensive review of FL-based heart disease detection methods. While several studies have explored the use of FL in healthcare and ML for detecting heart diseases, our paper specifically focuses on the application of FL to heart disease detection. We conducted a thorough review of the literature and identified a gap in the existing research, which prompted us to undertake this survey paper.

\section{Research Questions}
\label{rqs}
In this section we are set to address the following research questions based on the papers that we discussed and analyzed in the previous sections:
\begin{itemize}
\item \textbf{RQ1:} What are the different types of data used for detecting heart diseases using FL?
\item \textbf{RQ2:} What are the models proposed in different papers to detect heart diseases using FL?
\item \textbf{RQ3:} How does federated learning perform compared to traditional machine learning algorithms in detecting cardiovascular diseases?
\item  \textbf{RQ4:} What are the challenges and limitations of federated learning in detecting cardiovascular diseases, and how can they be addressed?
\end{itemize}

\subsection{RQ1: What are the different types of data used for detecting heart diseases using FL?}

\subsubsection{Electronic Health Records}
Electronic health records (EHRs) are the digital version of patients' medical records\cite{hipaa_gdpr_ref}. They contain information such as medical history, patient demographics, previous laboratory test results, and medication.  Hospitals are the major source of EHRs\cite{ehr_source}. Several publicly available datasets can be used for building and analyzing machine learning models. The authors of \cite{artificial_bee_colony_based_hdd} used the heart disease dataset of UCI Cleavland which has 303 health records and 76 attributes. The authors of \cite{predictive_models_from_electornic_health_records} used EHRs from Boston Medical Center that contains patient information about demographics, lab tests, vitals, tobacco use, emergency room visits, and past hospitalization records. The authors of \cite{fl_based_hdp} used the UCI Machine learning repository to obtain the EHRs with 14 attributes and 200 records.

\subsubsection{Electrocardiogram}
\begin{figure}[htp]
\includegraphics[scale=0.3]{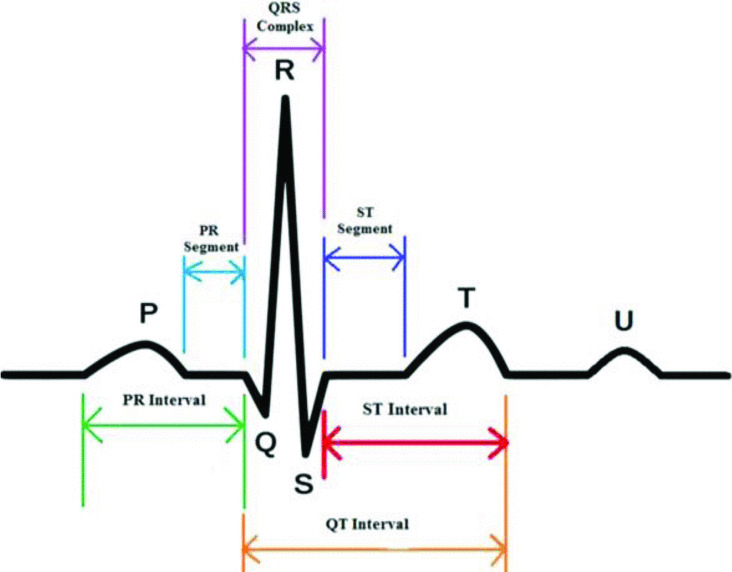}
\caption{Sample ECG signal\cite{ecg_image}}
\label{fig:ecg}
\end{figure}

Electrocardiogram (ECG) measures the electrical activity of the heart\cite{ecg_signals_ref}. It records the electric signals generated by the heart as it beats and pumps blood. These signals can be used to interpret the heart rate, detect any damaged parts of the heart or diagnose any heart conditions such as arrhythmia and CHFs\cite{ecg_classification_based_on_feature_alignment,ecg_ref}. Figure \ref{fig:ecg} is a sample ECG signal with different intervals labeled. The 'P' wave represents the contraction of the atrial muscle. The Q,R, and S represents the contraction of the ventricular muscle. The T wave represents the relaxation of the ventricles. The U wave is a small wave that follows the T wave and represents the repolarization of the papillary muscles. ECG signals should be preprocessed before they can be used for analysis. The authors of \cite{chf_based_on_unetpp} used the NSR-RR-interval and CHF-RR-interval databases in evaluating their proposed model. The raw ECG signals were sampled and digitized before feeding them to the model. The authors of \cite{fed_clustering} used the MIT-BIH arrhythmia database which contains 48 half-hour-long ECG reading from 47 subjects. Even here, the raw ECG signals are converted to digital signals by sampling at 360Hz. 

\subsubsection{Phonocardiogram}
\begin{figure}[htp]
\includegraphics[scale=2.8]{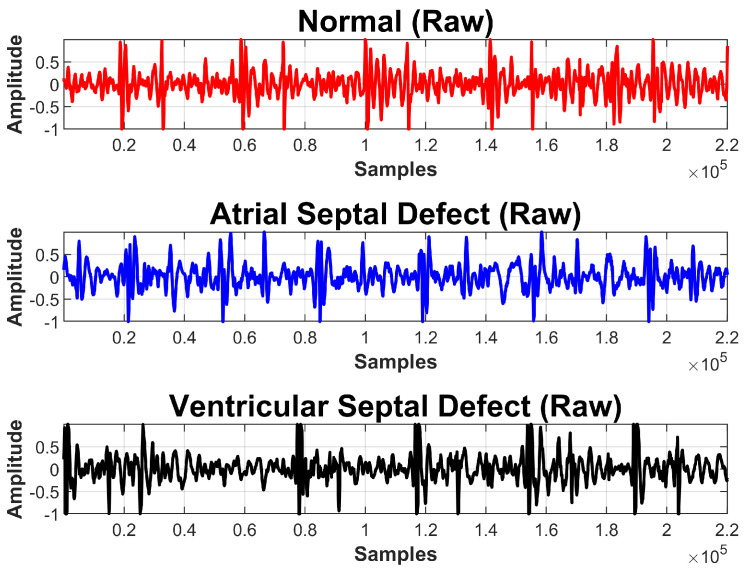}
\caption{Normal and abnormal PCG signals\cite{pcg_image}}
\label{fig:pcg}
\end{figure}

Phonocardiogram (PCG) is a reading of the sound of the heart, usually obtained by placing a microphone on the chest\cite{pcg_for_cad}. PCG can be used for diagnosing different pathological heart conditions\cite{phonocardiography_ref}. The analysis of PCG signals can provide information on the timing, intensity, and duration of heart sounds, which can be used to identify abnormalities in the heart. Sample PCG signals of normal and abnormal subjects is shown in Figure \ref{fig:pcg}. PCG data can be analyzed by ML models for detection of CVDs, for example, the authors of \cite{heart_sound_classification} used six different datasets with 3240 records in total obtained from 764 subjects. The PCG signals were converted to spectrograms for further processing using the wavelet analysis and neural networks.

\subsubsection{Echocardiogram}
Echocardiogram (echo) is a non-invasive imaging technique that uses ultrasound to produce images of the heart. It provides a real-time view of the heart's structures including the chambers, walls, valves, and blood flow\cite{echocardiogram_dl}. It is a useful diagnostic tool for detecting conditions such as CHFs and heart valve problems\cite{echocardiography_evaluation_ref}. It is time consuming processes to acquire high-quality images and the interpretation of the images is very subjective\cite{echo_ref}. However, the application of machine learning techniques can potentially alleviate these issues. The authors of \cite{hcm_with_ecg_echo}  collected echo data from 4 different institutions, identified the cases, labeled the data, and standardized the videos to 30 frames per second and resized to a square of size 299x299. This data is later fed to an FL-based 3-dimensional convolutional neural network.

\subsubsection{Cardiac Magnetic Resonance}
\begin{figure}[htp]
\includegraphics[scale=0.69]{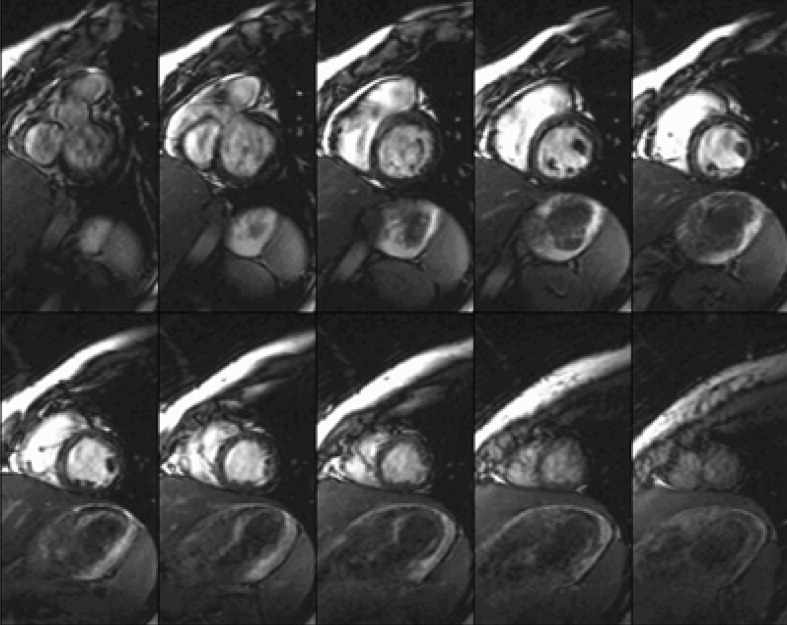}
\caption{Sample CMR image\cite{cmr_img_ref}}
\label{fig:cmr}
\end{figure}

Cardiac magnetic resonance (CMR) is an imaging technique used to image the heart and its surrounding structures. CMR uses a strong magnetic field, and electromagnetic pulses to generate high-resolution images of the heart and blood vessels\cite{cmr_ref}. A sample CMR image is shown in Figure \ref{fig:cmr}. It can be used to diagnose CHF, cardiomyopathy, and heart attacks. The authors of \cite{multi_center_imaging_diagnostics} used a combination of M\&M and ACDC datasets for the experiments. Contours were drawn onto the ACDC dataset by experts which are used for training the model.

\begin{table*}[htbp]
\caption{Summary of different data sources for heart disease diagnosis}
\begin{center}
\begin{tabular}{ | p{2.2cm} | c | p{3.5cm} | c | p{3.5cm} | c | }
\hline
\textbf{Data source} & \textbf{Abbreviation}& \textbf{Description}& \textbf{Type of data}& \textbf{Advantages}& \textbf{References}\\
\hline
Electrocardiogram& ECG& Measures the electrical activity of the heart& Electric signals& Non-invasive, widely available, low cost& \cite{ecg_ref}\cite{ecg_classification_based_on_feature_alignment}\\
\hline
Phonocardiogram& PCG& Records the sounds produced by the heart and blood vessels& Audio signals& Non-invasive, low cost, can detect some heart diseases earlier than other methods& \cite{pcg_image}\cite{heart_sound_classification}\\
\hline
Echocardiography& Echo& Uses ultrasound to visualize the heart's structure and function& Image data& Non-invasive, widely available, high diagnostic accuracy for many heart diseases& \cite{echocardiogram_dl}\cite{hcm_with_ecg_echo} \\
\hline
Cardiac Magnetic Resonance Imaging& CMR& Uses strong magnetic fields to produce detailed images of the heart and blood vessels& Image data& High spatial resolution, can provide detailed information about heart function and tissue characterization& \cite{hcm_with_ecg_echo}\cite{echocardiogram_dl}\\
\hline
Electronic Health Records& EHR& Comprehensive records of a patient's medical history, including diagnoses, medications, and laboratory results& Digital records& Large amount of data available for analysis, can provide a longitudinal view of disease progression& \cite{ehr_source}\\
\hline
\end{tabular}
\label{tab2}
\end{center}
\end{table*}

\subsection{RQ2: What are the models proposed in different papers to detect heart diseases using FL?}

In this section, we categorize the existing literature on Federated Learning for diagnosing heart diseases based on the models proposed.

\subsubsection{SVM with FedAvg}
Kavitha et al.\cite{fl_based_hdp} proposed and compared FL-based heart failure prediction models using Logistic Regression and Support Vector Machine (SVM). The federated learning algorithm used to build the global model was FedAvg. The models were trained to classify based on electronic health records. The experiment had five clients with twenty EHRs at each client. The logistic regression model achieved an accuracy of 82.38\% and SVM achieved an accuracy of 92.3\%. The results show that SVM performed better than Logistic Regression. Compared with the centralized SVM accuracy, FL SVM's accuracy was lower, but it was still in the acceptable range. Kavitha et al.\cite{fl_based_hdp} also compared the FL-based SVM accuracy with other centralized ML models that use EHRs as the input, and it shows that the FL-based SVM outperformed the existing centralized models.

\subsubsection{Cluster Primal Dual Splitting}
Theodora et al.\cite{predictive_models_from_electornic_health_records} proposed a method cluster Primal Dual Splitting to solve sparse SVM problem in a decentralized environment. It is based on the primal-dual splitting method and the idea of clustering, where the data is divided into several clusters, and each cluster is processed independently. The algorithm alternates between updating the primal variables associated with each cluster and the dual variables associated with the whole problem. The dataset used for the evaluation of the proposed algorithm was from the Boston Medical Center which contained EHRs of various patients belonging to different demographics. This data was used to predict the hospitalization of a patient with a given medical history due to heart disease. The results show that the proposed algorithm scored a better AUC score when compared to the existing methods. It converged faster that other existing algorithms. It also scored a lower computational cost which is ideal to run this algorithm at the edge devices with low computational capacity. 

\subsubsection{CNN with FedAvg}
\begin{figure}[htp]
\includegraphics[scale=0.3]{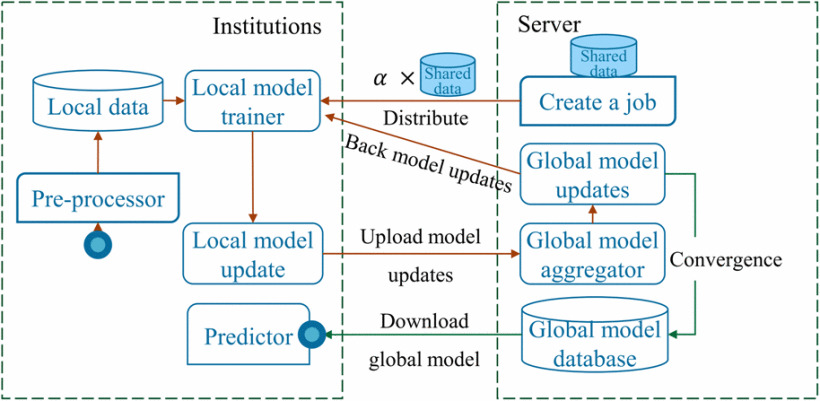}
\caption{FedAvg algorithm architecture for PCG classification\cite{heart_sound_classification}}
\label{fig:heart_sound_arch}
\end{figure}

 Wanyong et al.\cite{heart_sound_classification} compared the performance of two CNNs with different architectures and a multi-layer perceptron in a federated learning setup. The federated algorithm used was FedAvg. The models were trained to classify phonocardiogram data into normal or abnormal records. The architecture of the proposed model is shown in Figure \ref{fig:heart_sound_arch}. The local model training is triggered by the server. The local model updates are sent to the global model aggregator which updates the global model. This updated global model is later sent to the institutions as back model updates. The FL-based models were compared to their corresponding centralized versions. The performance of the models was evaluated with both IID and non-IID data distributions. The FL-based CNN has achieved an accuracy of 72.1\% when using IID data and 62.0\% when using non-IID data. The Fed-MLP model has shown poor performance and achieved an accuracy of 52.6\% with IID data and 45.3\% with non-IID data. In the case of centralized models, CNN based model has achieved an accuracy of 76.2\% and MLP based model achieved 57.9\%. Note that the lower accuracy values were due to the training of the model with limited data. The results show that the FL-based models perform better than the centralized models when dealing with low data volumes.

\subsubsection{Federated Matched Averaging with Modified Artificial Bee Colony}
 Muhammad et al.\cite{artificial_bee_colony_based_hdd} proposed a framework for diagnosing cardiovascular diseases with Federated Matched Averaging (FedMA) as the FL algorithm and Modified Artificial Bee Colony (M-ABC) as the optimizer. This proposed framework was evaluated and compared with different state-of-the-art FL frameworks. The experiment consisted of five nodes and one central server. The dataset used was the heart disease dataset of UCI Cleveland. The results show that the accuracy of the proposed framework is superior to the state-of-the-art FedMA with PSO optimizer, FedMA, and FedAvg. The convergence was also faster, and high accuracies were achieved with fewer communication rounds. When comparing the effect of the number of local epochs on the performance, all the existing algorithms without an optimizer showed a steep decline in performance after a certain number of local epochs. FedMA with PSO showed a slight decline in performance, but FedMA with M-ABC showed consistent performance after a certain number of local epochs. This implies that the local models can be trained as long as they want without affecting the global model's performance. When comparing the effect of accuracy on communication bandwidth, the proposed framework achieved higher accuracies with less communication volume than the other algorithms tested in this experiment. Overall, the proposed framework performed better than all the existing frameworks based on the evaluated metrics, providing a superior FL model to diagnose cardiovascular diseases accurately. However, the performance of this framework with more nodes and a larger volume of data still needs to be evaluated.

\subsubsection{Async-FL Model for Arrhythmia Detection}
\begin{figure}[htp]
\includegraphics[scale=0.3]{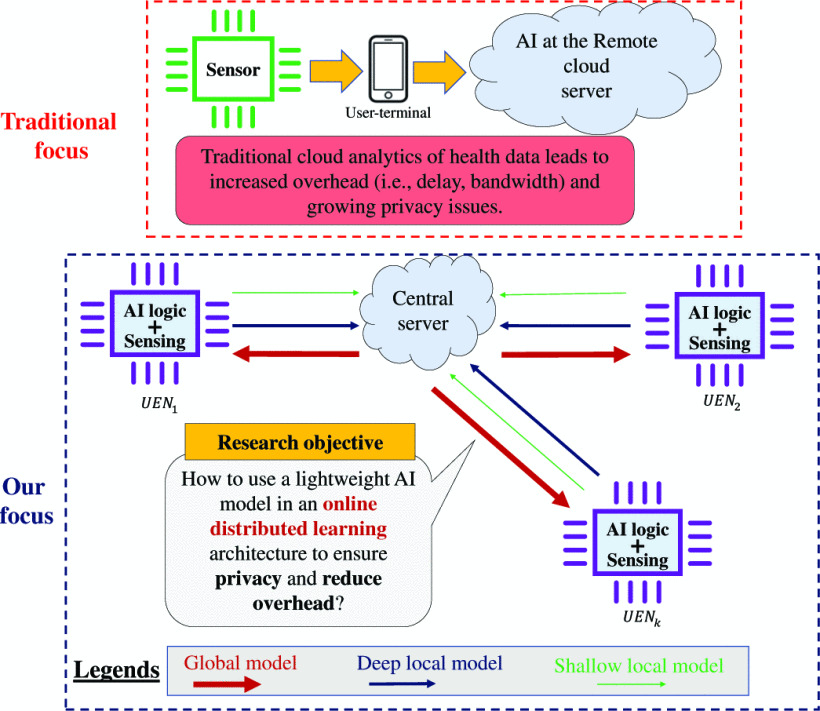}
\caption{Async FL-based ECG classification architecture\cite{async_ecg_arrhythmia}}
\label{fig:async_ecg}
\end{figure}

 Sadman et al. \cite{async_ecg_arrhythmia} proposed an FL-based asynchronous machine-learning model to detect cardiac arrhythmia from ECG signals at the edge devices. Later the performance of the async model was evaluated and compared with the performance of the sync model. The architecture of the model is shown in Figure \ref{fig:async_ecg}. Two algorithms were developed one that runs on the cloud and one that runs on edge devices. The algorithm that runs on the cloud takes the weights of the local models as the input and generates a global model as the output. The algorithm that runs on the edge devices will handle sending the local parameters to the cloud server asynchronously. The updates are sent to the server after completing a predetermined number of time rounds. The datasets used for evaluating the performance of the proposed model were the MIT-BIH Supraventricular Arrhythmia database, the MIT-BIH Arrhythmia database, INCART 12-lead Arrhythmia database, and the Sudden Cardiac Death Holter database. The heartbeats were classified into 4 classes\cite{prediction_of_coronary_artery_calcification_scores}, and the classifier used was a 1-D convolutional neural network. The evaluation metrics used were accuracy, weighted precision, weighted recall, and the AUC score. The number of edge nodes used in the experiment was 2, 4, 6, 8, and 10. The results show that the sync-FL model has shown better accuracy initially but both the models achieved almost the same accuracy of around 95\% after the completion of a few communication rounds. The sync-FL performed better with a lower number of nodes, and the async-FL performed better with more nodes. This implies that the async-FL can scale better than sync-FL. When the time taken for computation and memory used were compared, the async-FL model took shorter computational times, and the memory used was much lower than that of the sync-FL model. Though the accuracy was similar, the network efficiency, lower computational times, and lower memory requirements make async-FL superior to sync-FL.

\subsubsection{U-Net++}
\begin{figure}[htp]
\includegraphics[scale=0.38]{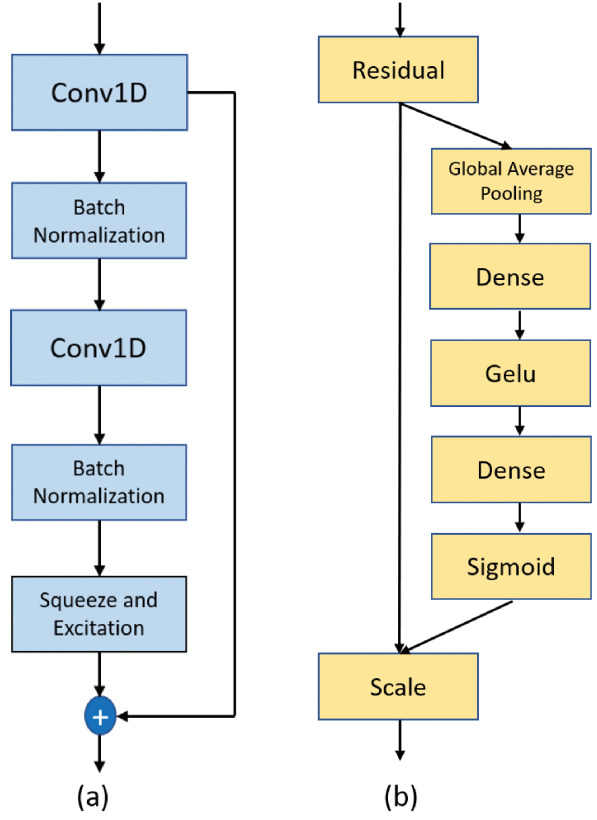}
\caption{Architecture of (a) convolutional unit and (b) SE block\cite{chf_based_on_unetpp}}
\label{fig:unetpp_arch}
\end{figure}

Liang et al.\cite{chf_based_on_unetpp} proposed an FL-based classification model that uses U-shaped networks to classify patients with CHF from ECG signals. The architecture of the model included two subnetworks, one for feature extraction and the other for classification. The architecture of the proposed model is shown in Figure \ref{fig:unetpp_arch}. The feature extraction subnetwork was composed of an encoder, a decoder, and a skip connection, while the classification subnetwork consisted of various components such as general convolutional layers, batch normalization, squeeze, and excitation block, global average pooling layers, and fully connected dense layers. The datasets used for evaluating the performance metrics of the proposed model are the NSR-RR-interval database and the CHF-RR-interval database. The NSR-RR-interval database has 54 normal sinus rhythm (NSR) ECG readings, and the CHF-RR-interval database has 29 CHF ECG readings. The number of nodes used was two to four. The performance of the FL model was evaluated over 230 communication rounds. The key metrics evaluated were accuracy, recall, precision, and F1-score. The centralized model converged after 18 epochs with a testing accuracy of 89.93\%, whereas the FL model converged after 50 communication rounds with a testing accuracy of 87.50\%. The proposed model outperformed all the existing CHF detection models in terms of accuracy. Even after inducing noise in the input data to simulate real-world raw ECG, the proposed model slightly degraded performance with an accuracy of 89.38\%, which is still in the acceptable range. To evaluate the model's performance in a practical scenario where the patient can have other CVDs such as apnea and not CHF, the model was trained and tested with CHF RR segments and AF or apnea RR segments. In this scenario, the centralized model scored an accuracy of 90.17\%, and the FL model scored an accuracy of 87.71\% with four clients. These experimental results suggest that the proposed model with U shaped network is superior to the existing models in detecting CHF in both centralized and FL scenarios, and the performance of the FL model is slightly lower than the centralized model.

\subsubsection{2D-CNN with Federated Aggregation}
Shinichi et al.\cite{hcm_with_ecg_echo} proposed an FL-based approach to train models to detect hypertrophic cardiomyopathy(HCM) based on ECG and echocardiogram data. A 2D convolutional neural network was used for classification and the federated algorithm used was federated aggregation. The data used for the evaluation of the model was from 4 medical institutions, the data from 3 medical institutions was used for training and the fourth is used for validation. The number of nodes used was 3. The results showed improved accuracy of the federated model when compared to the centralized model. This might be because the authors used data from 3 institutions to train the federated model and only one for the centralized model. So, the sensitivity scores were calculated and compared after training the FL-based model with a sub-sampled dataset. Even in this case, the FL managed to show a better AUROC score than the centralized model. The performance of the local FL models at each institution was better than the centralized model trained at a single institution.

\subsubsection{Federated Clustering}
\begin{figure}[htp]
\includegraphics[scale=0.30]{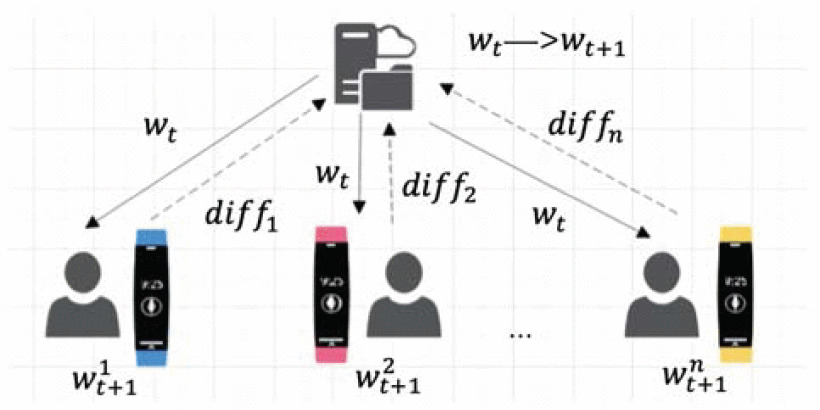}
\caption{Architecture diagram of FedClustering\cite{fed_clustering}}
\label{fig:fedcluster_arch}
\end{figure}

Daoquin et al.\cite{fed_clustering} proposed a novel federated algorithm FedCluster to address the highly unbalanced nature of the ECG data for the detection of cardiac arrhythmia. In this algorithm, the differential local model parameters sent to the server are first clustered. The averaging is first performed within the cluster and then in between the clusters. This results in increased weightage for the minority classes which counters the data imbalance. The architecture diagram of this model transitioning from $ t^{th} $ iteration to $ (t+1)^{th} $ iteration is shown in Figure \ref{fig:fedcluster_arch}.${w^n}_{t+1}$ is the weights of $n^{th}$ node in t+1 iteration and $ diff_n $ is the difference in weights of nth node from $t^{th}$ round to t+1 round. The performance of this method was tested using the MITBH arrhythmia database. The number of nodes used in the experiment was 24 and, the performance of the proposed FedCluster algorithm was compared to that of FedAvg. The results show that in the clients with less data imbalance, the improvement in prediction accuracy was low, but in the clients with the highest skewness in the data distribution, the improvement in the prediction accuracy was as high as 52\%. There was a significant improvement in accuracy for more than 30\% of the clients.

\begin{table*}[htbp]
\caption{Data Types and Algorithms Used in Various Papers}
\begin{center}
\begin{tabular}{ | c | p{4cm} | c | p{1.5cm} | c | p{5cm} | }
\hline
\textbf{Reference} & \textbf{Research Problem}& \textbf{Input type}& \textbf{Algorithm}& \textbf{Number of nodes}& \textbf{Limitations} \\
\hline
\cite{predictive_models_from_electornic_health_records}& Predict hospitalization due to heart diseases& EHRs& sSVM with cPDS& 4& May not converge quickly for certain classes of non-convex optimization problems \\
\hline
\cite{async_ecg_arrhythmia}& Develop a model to detect arrhythmia using async FL& ECG& Async FL with CNN& 2,4,6,8,10&  Scalability is not verified experimentally\\
\hline
\cite{heart_sound_classification}& Early diagnosis of CVDs using heart sound classification& PCG& FedAvg MLP, FedAvg CNN& 4& Performance is evaluated with limited data\\
\hline
\cite{artificial_bee_colony_based_hdd}&Improve the diagnosis method for the prediction of heart disease& EHRs& FedMA with M-ABC& 5& Scalability is not verified experimentally \\
\hline
\cite{multi_center_imaging_diagnostics}& To demonstrate the performance of FL in CMR diagnosis& CMR& FedAvg with 3D-CNN& 4& Proposed model is limited to binary classification\\
\hline
\cite{fed_clustering}& Propose a method that can handle highly unbalanced data& ECG& FedCluster& 24&\\
\hline
\cite{fl_based_hdp}& Propose a prediction model for heart diseases using FL& EHRs& FedAvg with LR and SVM& 5&  Effect of using non-IID data for training is not addressed\\
\hline
\cite{hcm_with_ecg_echo}& Propose FL model to detect HCM
& ECG, Echo& 2D-CNN with Federated aggregation& 3& \\
\hline
\cite{chf_based_on_unetpp}& Improved classification model for CHF& ECG& UNet++& 2,3,4& Dataset used for evaluation of the model is small and scalability is not discussed\\
\hline
\end{tabular}
\label{tab1}
\end{center}
\end{table*}

\subsection{RQ3: How does federated learning perform compared to traditional machine learning algorithms in detecting cardiovascular diseases?}
Traditional machine-learning algorithms require large volumes of data for training, which is difficult to obtain in healthcare due to privacy concerns\cite{privacy_in_ml_ref}. Federated learning can overcome this challenge by training the model at the edge devices and not sharing the raw data\cite{survey_on_fl_ref}. This gives FL access to data that is not accessible for traditional machine-learning algorithms. So, FL-based models show similar or superior performance when the data is scarce\cite{fl_ops_and_challenges_ref}.  In terms of dealing with data imbalance which is common in healthcare data, FL has shown better capability than the centralized models. FL can help mitigate the issues of dataset bias and overfitting because of data heterogeneity across various nodes. The results of the experiments performed in \cite{chf_based_on_unetpp}, the proposed FL model based on U-shaped networks for predicting CHF based on ECG signals showed prediction accuracy comparable to a centralized model while preserving privacy. This can also be observed in \cite{fl_based_hdp} where the proposed FL-based SVM model for detecting heart diseases from EHRs showed performance similar to the centralized SVM model. Although similar performances were observed, the potential for FL to use more data due to reduced privacy concerns and scale better than centralized versions is higher.

Most of the papers assume that the data is independent and identically distributed. But in the real world, the healthcare data collected from various institutions and edge devices is non-IID in nature\cite{fl_for_smart_healthcare_a_survey}. For example, a certain trait of a demographic causes a specific disease in a local health institution creating a bias in the data distribution which might be different for other institutions \cite{bias_due_to_demographics_ref}\cite{fl_in_medical_imaging}. The performance of FL-based models when dealing with non-IID data can be observed in the experimental results of \cite{heart_sound_classification}. The FL model was used to predict the chance of having a CVD using heart sounds. The data was heterogenous and non-identically distributed across 4 nodes. The results of this experiment showed the capability of FL in dealing with non-IID data. The lower accuracy values in this experiment were mainly due to the lack of training data. The comparisons are summarized in Table \ref{tab3}.

\begin{table*}[htbp]
\caption{Comparison of Centralized vs Federated Models for various model attributes}
\begin{center}
\begin{tabular}{ | c | p{6.3cm} | p{6.cm} | }
\hline
\textbf{Model Attribute} & \textbf{Centralized Model}& \textbf{Federated Model}\\
\hline
Volume of data required& More data is required& Less volume of data is sufficient\\
\hline
Data distribution handling capacity& Performance will be effected due to heterogeneous data& Can handle heterogeneous data\\
\hline
Model accuracy& Can score high accuracy& Slightly low accuracy than centralized models\\
\hline
Privacy& Pose higher risk to privacy due to data sharing& Improved privacy due to local training on devices\\
\hline
Computational requirement& Require single device with high computational capacity& Require more number of devices with less computational capacity\\
\hline
Communication overhead& Low& High\\
\hline
Robustness to device failure& Susceptible if central server goes down & Resilient to device failure\\
\hline
\end{tabular}
\label{tab3}
\end{center}
\end{table*}

\subsection{RQ4: What are the challenges and limitations of federated learning in detecting cardiovascular diseases, and how can they be addressed?}
Federated learning has the potential to be a useful tool for diagnosing cardiovascular diseases. But it also has some challenges and limitations, the primary challenge being the ability to handle non-IID data. Although FL can handle data distributions better than traditional machine learning techniques, skewed data can still decrease the performance of an FL-based model as demonstrated by \cite{fl_on_non_iid_data_a_survey}. Non-IID data could also cause communication inefficiencies leading to slower convergence of the model\cite{non_iid_convergence}. This issue was addressed by Daoquin et al.\cite{fed_clustering} by proposing a novel algorithm called FedCluster. It was used to detect cardiac arrhythmia from ECG signals. This algorithm has shown a significant improvement in the capability of the model in handling highly unbalanced data.

Data privacy and security are critical when dealing with highly sensitive data such as healthcare data. FL improves data privacy by preventing raw data transfer. But in some cases when using cross-silo architectures, where the data generated at the edge devices is sent to a local server of the institution, there is still a chance for compromising the data privacy and security\cite{safegaurding_cross_silo_fl}. This can be prevented by using different privacy-preserving techniques\cite{privacy_preserving_fl}\cite{safegaurding_cross_silo_fl}. Techniques as simple as adding noise to the model parameters when sending the updates to the central server can improve the security but it might affect the performance of the global model. Communication overhead is another challenge in FL\cite{communication_overhead_in_fl}. This overhead can occur at the network between edge devices and the local server in case of cross-silo and edge devices and the global server in case of a cross-device architecture. Sadman et al.\cite{async_ecg_arrhythmia} addressed this issue by proposing an async-FL model for detecting cardiac arrhythmia from ECG signals. The authors demonstrated the performance of async-FL which required less network bandwidth, low memory requirements, and improved scalability.

FL-based heart disease detection methods are still in the early stages of development, and there is currently no standard framework or protocol for developing and evaluating these methods\cite{standardization_in_fl_ref}. This lack of standardization can make it difficult to compare results across studies, limit the generalizability of findings, and hinder widespread adoption in clinical practice. Addressing this gap will require the development of standardized frameworks and protocols for developing and evaluating FL-based heart disease detection methods.

Another research gap is the training data used to develop FL-based heart disease detection models can be biased, which can lead to inaccurate predictions and limit the generalizability of the model. For example, if the training data only includes data from a particular demographic group, the resulting model may not generalize well to other groups\cite{data_generalization_in_fl}. Addressing this gap will require the development of more diverse and representative datasets, as well as the development of methods for detecting and mitigating bias in training data. FL-based methods can also be computationally expensive for edge devices, which can limit their scalability and usability in real-world settings. For example, if a model takes too long to train or make predictions, it may not be practical for use in a clinical setting. Addressing this gap will require the development of more efficient and effective FL-based algorithms that can handle large data volumes and provide real-time predictions.

Overall, addressing these research gaps will be critical to the development and adoption of FL-based heart disease detection methods in clinical practice. By developing more efficient and effective methods for developing and evaluating FL-based models, and by addressing issues related to data availability, bias, and computational complexity, researchers can help ensure that FL-based heart disease detection methods are effective, and widely adopted in clinical practice.

\section{Discussion}
\label{discussion}
In this survey paper, we reviewed several FL-based heart disease detection methods and evaluated their strengths and limitations. Our findings imply that FL-based approaches have the potential to improve the accuracy of heart disease detection while ensuring data privacy. FL solves privacy issues by training the model at the edge devices without disclosing raw data, allowing FL to access data that traditional machine-learning algorithms do not have access. Experiments with FL-based models for predicting congestive heart failure and detecting heart diseases have shown performance comparable to centralized models while preserving privacy\cite{chf_based_on_unetpp}. FL has the ability to leverage more data and scale more effectively than centralized versions, making it a compelling approach for healthcare applications.

However, FL is not without its challenges. Communication overhead and privacy concerns can still arise in some FL architectures, such as cross-silo architectures. To mitigate these challenges, privacy-preserving techniques such as adding noise to model parameters can be used\cite{adding_noise_to_params_in_fl}. The studies have shown that there was only a little impact of using this technique on the performance of the model\cite{differential_privacy_in_fl, privacy_preserving_fl, fl_privacy_ref}. Additionally, async-FL models have been proposed as a solution to reduce communication overhead and improve scalability\cite{async_ecg_arrhythmia}. Although FL can handle non-IID data better than centralized models, the performance of the FL model can still take a hit due to highly unbalanced data. We have discussed studies that focused on this data imbalance issue and reviewed algorithms such as FedClustering\cite{fed_clustering} that addressed this specific issue.

In addition, our study revealed significant limitations in current studies on FL-based cardiac disease detection. First, further research on processing extremely heterogeneous data, which is prevalent in a healthcare setting, is required. Second, there is a need for lightweight, computationally affordable algorithms that can run on low-powered edge devices. Third, additional research is needed to investigate FL's potential in conjunction with other developing technologies such as wearable devices and internet-of-things (IoT) devices. Finally, more research on the clinical efficacy and feasibility of FL-based cardiac disease detection in real-world healthcare settings is required.

\section{Conclusion}
In conclusion, FL has the ability to completely transform the diagnosis of CVDs. FL facilitates the pooling of patient data while ensuring data privacy and security by employing machine learning techniques and collaborative learning across many healthcare organizations. This method results in more precise CVD diagnosis, earlier identification, personalized therapy, and lower healthcare expenditures. Although FL is a relatively new technology, recent research has demonstrated that it has the potential to improve the accuracy and efficacy of CVD detection. The capacity to exchange data across several institutions, in particular, enables the analysis of a larger pool of patient data, which can increase the accuracy of CVD diagnoses and treatment programs

However, there are still challenges that must be addressed. One of the most difficult challenges is the ability to handle highly heterogeneous data collected across multiple institutions. It is critical to build more advanced FL algorithms that can account for the variability of patient data across institutions. Communication overheads and data security are further challenges that must be addressed. Nonetheless, the survey results demonstrate that FL has the potential to revolutionize CVD diagnosis and treatment. We feel that additional study is required to address these difficulties and fulfill FL's promise for CVD detection. Future research could focus on developing more efficient and effective FL algorithms, optimizing model aggregation techniques, and exploring new data sources for CVD detection. Overall, we hope that this survey paper serves as a valuable resource for researchers and practitioners working in the field of FL and CVD detection.


\end{document}